\documentclass[11pt]{article}

\usepackage[preprint]{acl}
\usepackage{booktabs} 
\usepackage{times}
\usepackage{latexsym}
\usepackage{amsmath}
\usepackage{amssymb}
\usepackage{bm}
\usepackage{mathtools}
\usepackage{amsthm}
\usepackage{xspace}
\usepackage{adjustbox}
\usepackage{multirow}
\usepackage{enumitem}
\usepackage[table]{xcolor} 
\usepackage[T1]{fontenc}
\usepackage{ulem} 

\newif\ifdraft
\drafttrue      

\usepackage[utf8]{inputenc}

\usepackage{microtype}

\usepackage{inconsolata}

\usepackage{graphicx}

\def\eqref#1{equation~\ref{#1}}

\def\1{\bm{1}}

\def\vg{{\bm{g}}}

\def\vk{{\bm{k}}}

\def\vq{{\bm{q}}}

\def\vv{{\bm{v}}}

\def\vy{{\bm{y}}}

\def\mK{{\bm{K}}}

\def\mV{{\bm{V}}}

\DeclareMathAlphabet{\mathsfit}{\encodingdefault}{\sfdefault}{m}{sl}
\SetMathAlphabet{\mathsfit}{bold}{\encodingdefault}{\sfdefault}{bx}{n}

\def\equationautorefname~#1\null{Eq.~(#1)\null}

\title{Augmenting Attention with Exponentially Decaying Memory Improves Query-Aware KV Sparsity}

\author{Xiuying Wei \\
  EPFL, CLAIRE \\
  Lausanne, Switzerland \\
  \texttt{xiuying.wei@epfl.ch} \\\And
  Caglar Gulcehre \\
  EPFL, CLAIRE \\
  Lausanne, Switzerland \\
  \texttt{caglar.gulcehre@epfl.ch} \\}

\begin{document}
\maketitle
\begin{abstract}
Efficient inference is critical for long-context language models, where attention computation and KV-cache access dominate the cost. Recent work, RAT+~\citep{ratplus}, introduces a recurrence-augmented attention backbone that enables flexible dilated attention at inference time. In this paper, we investigate whether this exponentially decaying memory can also improve existing query-aware sparse inference methods. Using representative methods including Quest, MoBA, and SnapKV, we show that RAT+ consistently improves accuracy over standard attention across sparse budgets on eight needle-in-a-haystack tasks. We validate these gains both on the released checkpoints from the RAT+ paper and on OLMo2-7B, which we continue pretraining with the added memory module for 10B tokens. Finally, we propose two hypotheses explaining why this memory module benefits query-aware sparse inference and design targeted experiments to support them. 
\end{abstract}

\section{Introduction}
Efficiency has become a major concern for attention modules~\citep{attention}, whose computation and memory costs scale quadratically with context length. 
Many works study sparse attention at inference time, ranging from local patterns~\citep{attention_sink} to dynamic query-aware patterns. 
Instead of attending to all tokens, some methods select critical blocks of tokens for lightweight computation, motivated by the sparsity of attention maps~\citep{moba,quest,twilight,infllm}. Others reduce the KV cache by identifying critical tokens for each input and retaining them for subsequent decoding~\citep{h2o,snapkv,ada-kv}.  While effective, these methods still show clear accuracy degradation under challenging settings, such as 98\% sparsity for Quest~\citep{quest, rectified_sparse_attention}, over 75\% KV cache storage reduction for SnapKV~\citep{snapkv,ada-kv}, and evaluations on harder tasks.

RAT+~\citep{ratplus} introduces a recurrence-augmented attention backbone that applies a exponentially decaying memory over KV states, enabling flexible-budget dilated attention at inference time. While RAT+ focuses on dilated inference, we investigate whether the same architecture can also improve existing query-aware sparsity methods with dynamic patterns. 

We evaluate three representative query-aware sparse inference methods from different efficiency categories: Quest~\citep{quest} for reducing decoding-time FLOPs, MoBA~\citep{moba} for reducing prefilling-time FLOPs, and SnapKV~\citep{snapkv} for reducing KV-cache storage. On needle-in-a-haystack (NIAH) tasks from the RULER benchmark~\citep{ruler}, these methods achieve substantially higher accuracy when applied to the RAT+ backbone rather than to the standard attention backbone. We validate this finding both on the released RAT+ checkpoints~\citep{ratplus} and on OLMo2-7B~\citep{olmo2}, which we continue pretraining with the additional memory module for only 10B tokens. For instance, SnapKV improves by 34.11 and 40.03 points on average across eight tasks under 1/4 and 1/8 budgets, respectively; on OLMo2-7B, Quest improves from 68.0 to 98.6 on MK-2 under a 1/16 budget, while MoBA improves from 53.6 to 94.8 on MK-3. 

Finally, we propose two hypotheses to explain these gains: exponentially decaying memory improves critical-token selection accuracy, and it provides an additional information path for selected candidates, thereby preserving more answer-related information. We carefully design experiments to demonstrate both hypotheses. Overall, we hope our work sheds new light on efficient inference: besides focusing solely on improving downstream inference methods, we can also design upstream architectures that are inherently more capable with sparse inference. 

\section{Method}
\subsection{Exponentially decaying memory}
RAT+~\citep{ratplus} augments standard attention with a lightweight recurrence and uses additional optimization techniques to obtain an effective recurrence length of 64. We view this recurrence as an exponentially decaying memory. At each time step $t$, the value and key states are updated as 
\begin{equation} 
\label{eq:recurrence} 
\begin{aligned} 
\tilde{\vv}_{t} &= \vg_{t}\odot \tilde{\vv}_{t-1} + (1 - \vg_{t})\odot \vv_{t},\\ 
\tilde{\vk}_{t} &= \vg_{t}\odot \tilde{\vk}_{t-1} + (1 - \vg_{t})\odot \vk_{t}, 
\end{aligned} 
\end{equation}
where $\vg_t$ is an input-dependent gate. By controlling the effective memory length to 64, RAT+ allows this lightweight module to operate stably. It introduces only $\mathcal{O}(1)$ additional computation and storage per token, and can be implemented efficiently.

While this was originally introduced to support flexible dilated attention at inference time, we use its dense version as a memory-augmented backbone. 
Given a sparse inference method $\mathcal{S}$, we compare two instantiations: applying $\mathcal{S}$ to the original KV states from a standard attention backbone, or to the memory-augmented states from RAT+:
\begin{equation}
\label{eq:sparse_compare}
\mathcal{S}(\mK,\mV)
\quad \text{vs.} \quad
\mathcal{S}(\tilde{\mK},\tilde{\mV}).
\end{equation}
We instantiate $\mathcal{S}$ with three representative sparse inference variants below.  
\subsection{Sparse Inference Instantiations}
We formulate query-aware sparse inference as a two-stage procedure: selecting critical KV candidates and computing attention over the selected states. 
Given the augmented KV states $(\tilde{\mK},\tilde{\mV})$ and a selection query $\vq^{\mathrm{sel}}$, the selected index set is
\begin{equation}
\label{eq:selection}
\mathcal{I}
=
\operatorname{TopK}_{i}
\; s\!\left(\vq^{\mathrm{sel}}, \rho(\tilde{\mK}_i)\right),
\end{equation}
where $s$ is a scoring function and $\rho(\tilde{\mK}_i)$ is the representation of the candidate $i$. 
Attention is then computed only over the selected states:
\begin{equation}
\label{eq:evaluation}
\vy_t
=
f\left(\vq_t \tilde{\mK}_{\mathcal{I}}^\top\right)
\tilde{\mV}_{\mathcal{I}}.
\end{equation} 
We consider three representative methods, as shown in \autoref{tab:sparse_methods}, each targeting a different aspect of inference efficiency.  Quest for decoding and MoBA for prefilling operate at the block level, using block representatives as $\rho$ and the current query $\vq_t$ as $\vq^{\mathrm{sel}}$.  SnapKV instantiates the candidates as individual KV tokens and uses the last prefilling tokens as an observation window for $\vq^{\mathrm{sel}}$.  By discarding unselected entries, SnapKV reduces both KV cache size and FLOPs in subsequent decoding. 

\begin{table}[t]
\centering
\small
\caption{Our evaluated sparse patterns with RAT+.}
\begin{adjustbox}{max width=\linewidth}
\begin{tabular}{llll}
\toprule
\bf Method & \bf Unit & \bf FLOPs reduction & \bf KV storage reduction\\
\midrule
Quest  & Block & Decode & No \\
MoBA   & Block & Prefill & No \\
SnapKV & Token & Decode & Yes \\
\bottomrule
\end{tabular}
\end{adjustbox}
\label{tab:sparse_methods}
\end{table}
\begin{figure*}[htbp!]
    \centering
    \includegraphics[width=\linewidth]{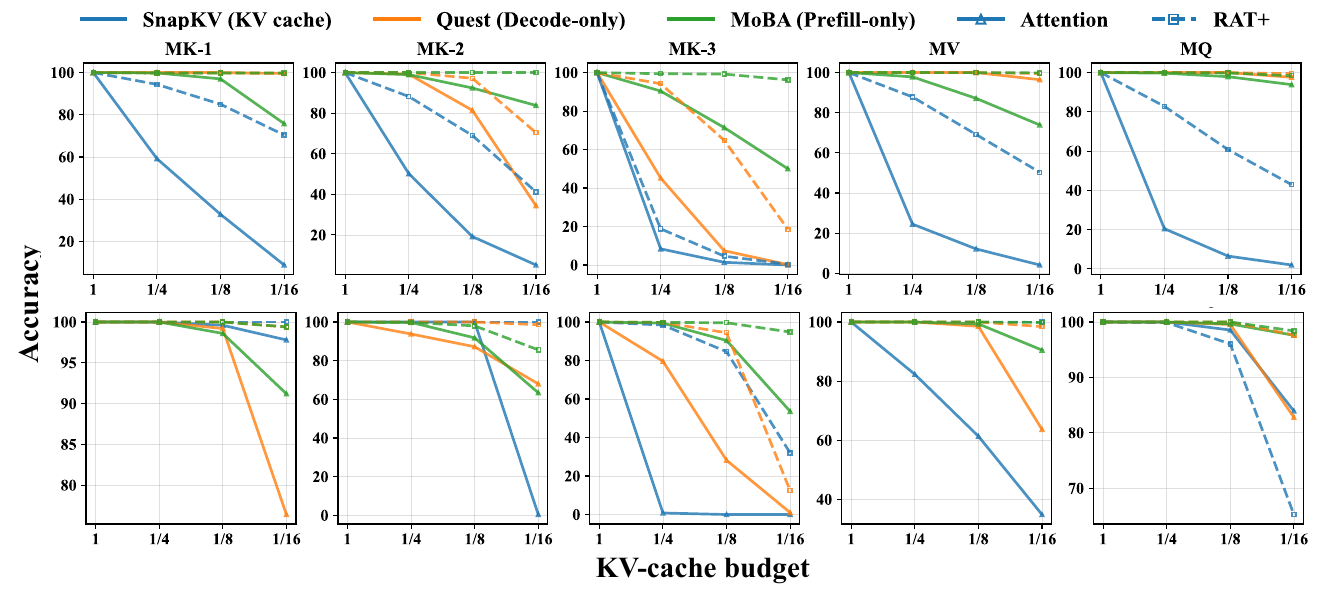}
    \caption{
    Performance on eight NIAH tasks of two types of checkpoint. Due to space limits, the three NIAH-Single tasks and additional challenging settings are reported in \autoref{tab:llama_7b} and \autoref{tab:olmo_7b}.  \textbf{Top row}: evaluation on the 7B checkpoints from \citet{ratplus}, where standard attention and RAT+ are pretrained from scratch on 100B tokens.  \textbf{Bottom row}: evaluation on OLMo2-7B~\citep{olmo2}, which is pretrained on trillions of tokens; the corresponding RAT+ backbone is obtained by 10B-token continued pretraining with the additional exponentially decaying memory module.  The KV budget denotes per-query KV access; Quest and MoBA still store the full KV cache. Some curves overlap due to near-perfect accuracy.
    }
\label{fig:llama_7b}
\end{figure*}
\subsection{Analyses}
We analyze the effect of this memory module from both candidate selection and sparse computation.

\paragraph{H1: Exponentially decaying memory improves critical-token selection.}
This hypothesis corresponds to the selection stage in \autoref{eq:selection}. 
We hypothesize that memory-augmented states increase the likelihood that answer-relevant KV candidates are included in the selected blocks or tokens. 
To test this, we use a strict head-level hit rate aligned with exact-match retrieval. 
Let $g_t$ denote the golden answer position required at decoding step $t$, and let $\mathcal{I}_{h,t}$ denote the candidates selected by head $h$ at that step. 
We define
\[
\mathrm{Hit}_h =
\mathbf{1}
\left[
\forall t \in \{1,\ldots,T\},\;
g_t \in \mathcal{I}_{h,t}
\right],
\]
where a head receives a score of 1 only if its selected candidates contain the required golden answer position at every decoding step.

\paragraph{H2: Exponentially decaying memory serves as an additional information path for selected candidates.}
This hypothesis corresponds to the sparse computation stage in \autoref{eq:evaluation}. 
We hypothesize that memory-augmented states carry more answer-relevant information, even when the selected blocks or tokens miss the exact answer positions.  This is motivated by \citet{transformer-xl}, where recurrence propagates information across distant positions.  To isolate this effect from selection quality, we replace the original selector in \autoref{eq:selection} with a random selector:
\[
\mathcal{I}^{\mathrm{rand}} \sim \mathrm{Unif}\left(
\left\{\mathcal{I}\subseteq \{1,\ldots,M\}
:\ |\mathcal{I}|=K\right\}\right),
\]
where $M$ is the number of available candidates and $K$ is the original sparsity budget.
We repeat each randomized evaluation with five seeds and report the averaged performance.

\section{Experiments}
\subsection{Setup}
We consider two model settings: (1) the 7B checkpoints released by \citet{ratplus}, where both standard attention and RAT+ are pretrained from scratch on 100B tokens; and (2) OLMo2-7B~\citep{olmo2}, a standard-attention model pretrained on trillions of tokens, for which we obtain a RAT+ counterpart by continuing pretraining for 10B tokens with the additional memory module. Note that the exponentially decaying memory receives only limited training compared with the original model parameters due to our resource limits; our later used  supervised fine-tuning can enable fairer comparisons. We apply Quest, MoBA, and SnapKV following \autoref{tab:sparse_methods} to all layers under $1/4$, $1/8$, and $1/16$ sparse budgets, which specify the fraction of activated KV entries per query and approximately correspond to FLOPs reduction. Evaluation is conducted on the eight needle-in-a-haystack tasks from RULER~\citep{ruler}, which require models to retrieve inserted numbers or UUID-like strings from 4K context window with distractors. We apply lightweight one-stage supervised fine-tuning before evaluation to reduce prompt-following sensitivity and focus on architectural differences, especially for the pretrained-only checkpoints. Details are provided in \autoref{sec:implementation_details}. 


\subsection{Main results}
\autoref{fig:llama_7b} show that RAT+ consistently outperforms standard attention in sparse inference methods and KV budgets. This trend not only holds for checkpoints in \citet{ratplus}, but also the OLMo2-7B, for which we additionally train the exponentially decaying memory with 10B tokens. This suggests that the benefit also transfers to existing industry-scale pretrained models. In the $1/16$ budget, RAT+ improves SnapKV in S-1 from 39.2 to 84.2 in \autoref{tab:llama_7b}. For Quest, standard attention is already near-perfect on simple NIAH-Single tasks, but RAT+ substantially improves harder multi-key settings, e.g., MK-2 from 76.4 to 99.4 and MK-3 from 68.0 to 98.6 on the OLMo checkpoints. We further evaluate more challenging settings for Quest and MoBA in \autoref{tab:llama_7b}, \autoref{tab:olmo_7b} and observe stronger gains.


\begin{figure}[htbp!]
    \centering
    \includegraphics[width=\linewidth]{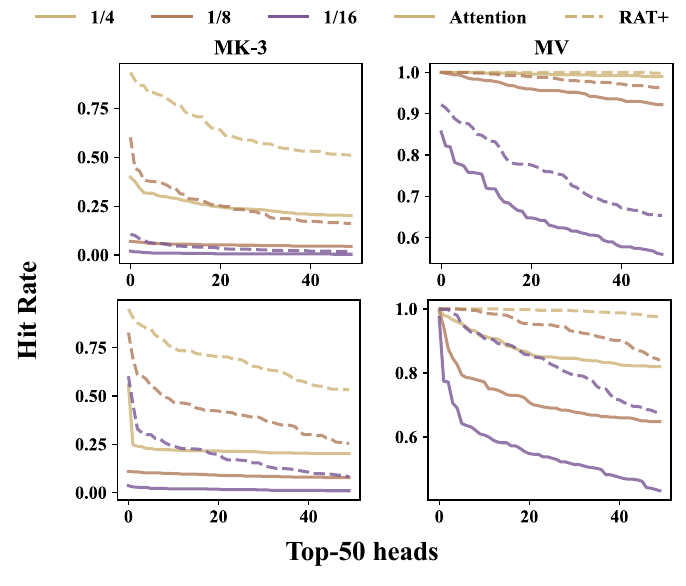}
    \caption{
    H1: Head-level hit-rate distributions for the top-50 heads ranked by hit rate.
    We mainly analyze Quest (\textbf{top row}) and SnapKV (\textbf{bottom row}) in H1,
    because their token selection during decoding is more directly aligned with
    individual generated answer tokens. Results are reported on the released 7B
    checkpoints from \citet{ratplus}.
    Additional results are provided in \autoref{fig:hit_rate_snapkv}, \autoref{fig:hit_rate_quest}.
    }
    \label{fig:hit_rate}
\end{figure}

\subsection{Analyses}
\paragraph{H1 results.}
Since NIAH tasks are retrieval-heavy, the final answer often depends on a few retrieval heads~\citep{wu2024retrieval}.  We therefore rank heads by hit rate and compare the top heads between standard attention and RAT+.  For SnapKV, the candidate set is fixed after prefilling, so the hit rate directly measures whether answer-relevant tokens are retained in the reduced KV cache. For Quest, which selects blocks at each decoding step, the hit rate reflects the final result of step-wise dynamic selection.  MoBA performs selection only during prefilling, so its selected blocks are less directly aligned with each generated answer token; we therefore focus the H1 analysis on Quest and SnapKV.  As shown in \autoref{fig:hit_rate}, RAT+ consistently yields higher hit rates, especially in settings with larger accuracy gains. 

\paragraph{H2 results.}
\autoref{tab:random_selector} supports our second hypothesis. 
We replace the designed selector in \autoref{eq:selection} with a random selector and evaluate all methods with the budget of $1/4$. 
RAT+ consistently outperforms standard attention, e.g., improving Quest from 53.4 to 84.9 on MK-1 and from 39.6 to 74.4 on MV. 
Since the selected candidates are random, these gains cannot be explained by more accurate selection. 
Instead, they suggest that memory-augmented selected states carry more answer-relevant information, leading to higher accuracy and closing much of the gap to the original selector even when the candidate set is imperfect.
\begin{table}[htbp!]
\centering
\caption{Random selector performance with \(1/4\) budgets reported on mean ± std with five seeds on OLMo2-7B checkpoints. See \autoref{tab:random_selector_full} for full performance.}
\label{tab:random_selector}
\resizebox{\linewidth}{!}{
\begin{tabular}{lll ccc ccc cc}
\toprule
 \bf Backbone & \bf MK-1 & \bf MK-2 & \bf MK-3 & \bf MV & \bf MQ\\
\midrule
\bf Quest \\
\hspace{0.5em}Attention & 53.4 $\pm$ 0.67 & 63.3 $\pm$ 2.42 & 0.1 $\pm$ 0.10 & 39.6 $\pm$ 0.61 & 56.9 $\pm$ 1.20 \\
\hspace{0.5em}RAT+ & \textbf{84.9 $\pm$ 0.73} &\textbf{ 91.5 $\pm$ 0.90} & \textbf{10.1 $\pm$ 1.16} & \textbf{74.4 $\pm$ 0.77} & \textbf{79.9 $\pm$ 0.31} \\
\midrule
\bf MoBA \\
\hspace{0.5em}Attention & 73.6 $\pm$ 1.15	& 77.4 $\pm$ 2.97	& 65.1 $\pm$ 1.42	& 89.2 $\pm$ 0.65	& 94.7 $\pm$ 0.38 \\
\hspace{0.5em}RAT+ &  \textbf{94.8 $\pm$ 0.85} & \textbf{97.1 $\pm$ 0.81} & \textbf{85.7 $\pm$ 0.64} & \textbf{92.3 $\pm$ 0.32} & \textbf{98.9 $\pm$ 0.21} \\
\bottomrule
\end{tabular}
}
\end{table}
\section{Related work}
Motivated by the sparsity of attention maps, many works study sparse attention for efficient inference.  Static methods use fixed patterns such as local-window attention~\citep{attention_sink}, while recent query-aware methods dynamically select important tokens or blocks for each query or sample.  Representative examples include Quest~\citep{quest}, which uses min/max block statistics, MoBA~\citep{moba}, which uses mean-pooled block representations, and Twilight~\citep{twilight}, which replaces selection of the top-$k$ with selection of the top-$p$. 
These methods reduce per-query computation and KV access, but usually keep the full KV cache. A separate line of work reduces the KV cache storage by retaining only important entries for later decoding. 
H2O~\citep{h2o} identifies heavy-hitter tokens, while SnapKV~\citep{snapkv} maintains important KV entries based on attention scores from an observation window. 
Unlike these downstream sparse inference designs, we study whether an upstream memory-augmented attention backbone can make existing query-aware sparse methods more accurate. 
The backbone we used \citet{ratplus} which was originally proposed for flexible dilated attention; here, we conduct a thorough study of whether this backbone without dilation can benefit existing query-aware sparse methods.
\section{Conclusion}
We investigate whether exponentially decaying memory improves query-aware KV sparsity. Across three representative methods with different efficiency goals, RAT+ substantially improves sparse inference accuracy over standard attention across inference budgets. Our analyses further explain the sources of these gains. These results suggest a broader view of efficient inference: it can also be improved by building more capable upstream architectures. 
\section*{Limitations}
Our experiments mainly evaluate two types of 7B checkpoints with a maximum sequence length of 4K. Therefore, we do not evaluate longer context lengths. Moreover, we mainly focus on needle-in-a-haystack tasks from the RULER benchmark. Their synthetic property allows us to generate training splits and tune both the original checkpoints and the newly added parameters, which are not trained as sufficiently as the original parameters in OLMo2-7B, providing a fairer comparison between backbones. Although the RULER tasks include eight variants with different retrieval patterns and are suitable for controlled continued-pretraining experiments, we do not evaluate other downstream tasks. Finally, we study three representative query-aware sparse inference methods: Quest, MoBA, and SnapKV. While they cover different efficiency goals, we do not evaluate other sparse inference methods. 
\section*{Acknowledgments}
Xiuying Wei’s work is supported by the DVPS project, funded by the European Union’s Horizon Europe Framework. We also sincerely thank the Swiss AI Initiative and the Swiss National Supercomputing Centre (CSCS) for supporting the computation through grants under project IDs a109. We extend our appreciation to Karin Getaz for administrative support. 
\bibliography{custom}
\clearpage
\appendix
\section{Appendix}
\label{sec:appendix}
\subsection{Implementation details}
\label{sec:implementation_details}
\paragraph{Models.}
We first evaluate the 7B checkpoints released by \citet{ratplus}, which include a standard attention backbone and a RAT+ backbone trained from scratch for 100B tokens. We further evaluate on OLMo2-7B, a pretrained model trained on up to 5T tokens. To obtain its RAT+ counterpart, we continue pretraining OLMo2-7B for 10B tokens from \citet{fineweb_edu} with added exponentially decaying memory. We use a learning rate of $6.0\times10^{-4}$ for the memory parameters and $6.0\times10^{-5}$ for the remaining parameters. The global batch size is 768, the sequence length is 4096, the warmup ratio is 2\%, and the weight decay is 0.1. Since the added parameters receive only limited additional training compared with the original model parameters due to our resource limitations, we further apply supervised fine-tuning on NIAH tasks to tune both backbones and make the results fairer.

\paragraph{Sparse inference settings.}
Quest~\citep{quest} is a KV-block selection method for decoding-time sparse attention. It partitions the KV cache into blocks and maintains lightweight block-level representatives constructed from the dimension-wise minimum and maximum key values. During decoding, each new query first interacts with these representatives to identify the most relevant top-$k$ blocks, and attention is then computed only over the KV entries within the selected blocks. Therefore, Quest reduces KV memory access and attention computation during decoding, while still retaining the full KV cache in memory. For Quest, we mainly use a block size of 64 and vary the number of selected blocks from 16 to 4, corresponding to different KV budgets. Compared with the original block size of 16, a block size of 64 makes selection more challenging because it operates at a coarser granularity, but it also introduces fewer additional FLOPs and enables more contiguous memory access. We also evaluate an extreme setting with block size 16 and 4 selected blocks in \autoref{tab:llama_7b}.

MoBA~\citep{moba} is a block-sparse attention method mainly used in the prefilling stage. The original method applies continued pretraining, but it also claims as a training-free paradigm and does not introduce new parameters, so we directly use it at inference time. Similar to Quest, MoBA routes each query to a subset of relevant KV blocks. Unlike Quest, however, MoBA uses mean pooled block representations for routing instead of dimension-wise minimum and maximum statistics. We mainly use a block size of 64 in the main experiments and further evaluate a block size of 128, since larger block sizes are often more favorable for prefilling due to better parallel efficiency. For both Quest and MoBA, we always include the first block and local blocks following their original designs.

SnapKV~\citep{snapkv} is a KV-cache reduction method applied after prefilling. It uses an observation window near the end of the prompt to estimate the importance of previous KV positions based on their attention scores. Only the selected important KV entries are retained in the KV cache, while the remaining entries are discarded. During subsequent decoding, attention is computed only over the retained KV entries. Therefore, SnapKV reduces both KV-cache storage and decoding-time attention cost. For SnapKV, we follow the original paper and use an observation window of 64. We set the number of retained KV entries $K$ to 1024, 512, and 256, corresponding to $1/4$, $1/8$, and $1/16$ KV budgets. 

\paragraph{Tasks.}
We evaluate on needle-in-a-haystack tasks from the RULER benchmark~\citep{ruler} with 500 examples for each sub-tasks. These retrieval-heavy tasks require models to identify target keys and values, such as numbers or long UUID-like strings, from long contexts with background noise or similar key-value pairs. We use eight NIAH settings: S-1 retrieves a single number from irrelevant background noise; S-2 retrieves a single number from natural-text background; S-3 retrieves a single UUID string from a context containing similar needle-like distractors; MK-1 retrieves the number corresponding to a target key among multiple key-value pairs with different keys; MK-2 increases the number of distracting key-value pairs; MK-3 further increases the distractor complexity and uses UUID strings as values; MQ requires retrieving numbers for multiple queried keys; and MV requires retrieving multiple numbers associated with the same key.

To isolate architectural differences from prompt-following sensitivity, we apply a one-stage supervised fine-tuning procedure before evaluation. We synthesize an additional training split from the same task generator using a different random seed, resulting in approximately 7M training tokens. This step helps the models better follow task-specific prompts, since the checkpoints from \citet{ratplus} are not instruction-tuned. For supervised fine-tuning, we use a learning rate of $1.0\times10^{-5}$ and a global batch size of 32 for all models. 

\paragraph{Experiment budgets}
For continued pretraining, we use 16 NVIDIA GH200 120GB nodes, each with 4 GPUs, to train the 7B model for 10B tokens. This takes approximately 10 hours. For supervised fine-tuning, we use 4 nodes for approximately 2 hours. Evaluation on all eight tasks takes about 20 minutes per setting.

\subsection{Supplementary experiments}
\begin{figure*}[t]
    \centering
    \includegraphics[width=\linewidth]{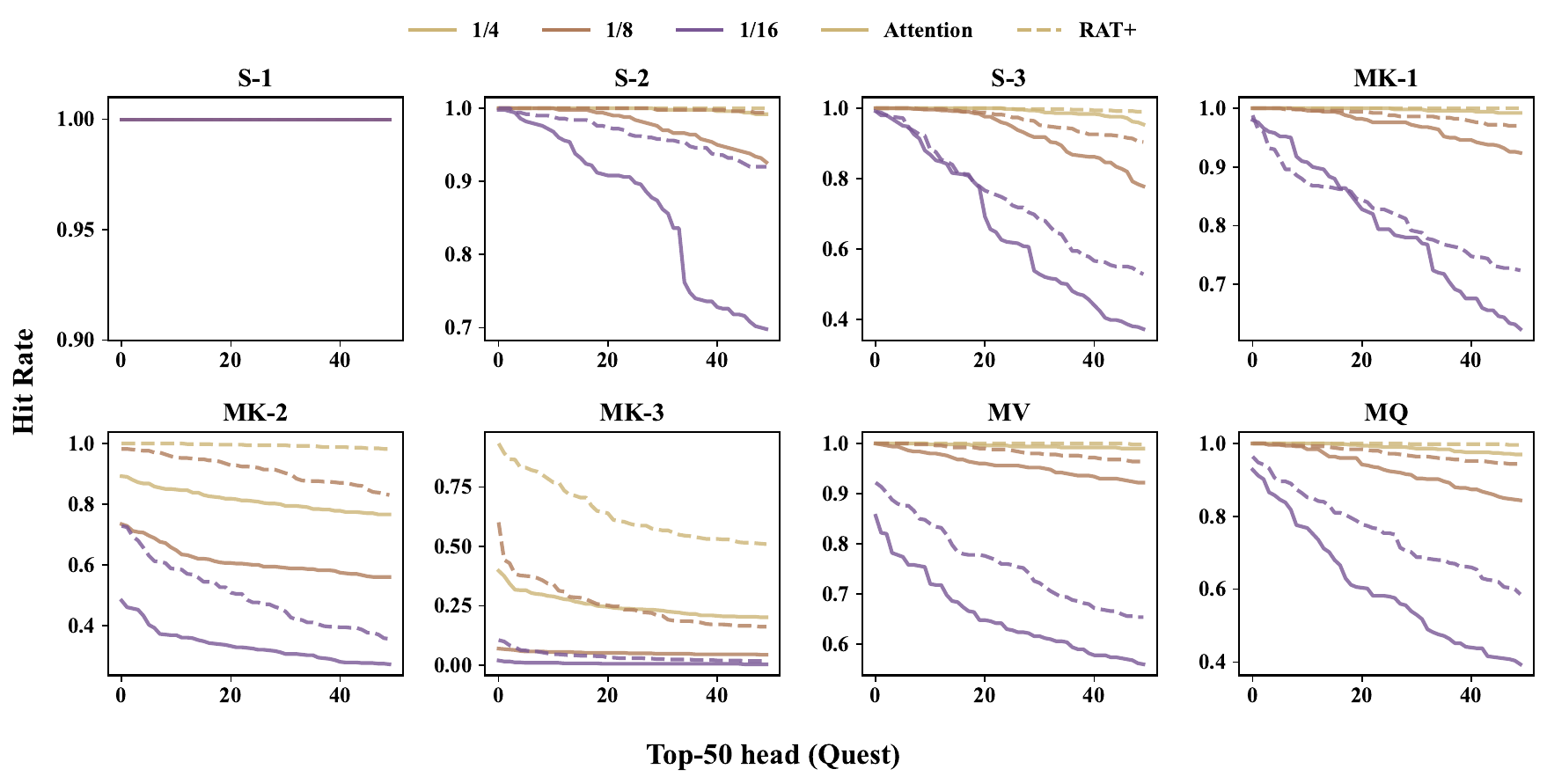}
    \caption{Head-level hit-rate distributions for the top-50 heads ranked by hit rate for Quest.}
    \label{fig:hit_rate_quest}
\end{figure*}

\begin{figure*}[t]
    \centering
    \includegraphics[width=\linewidth]{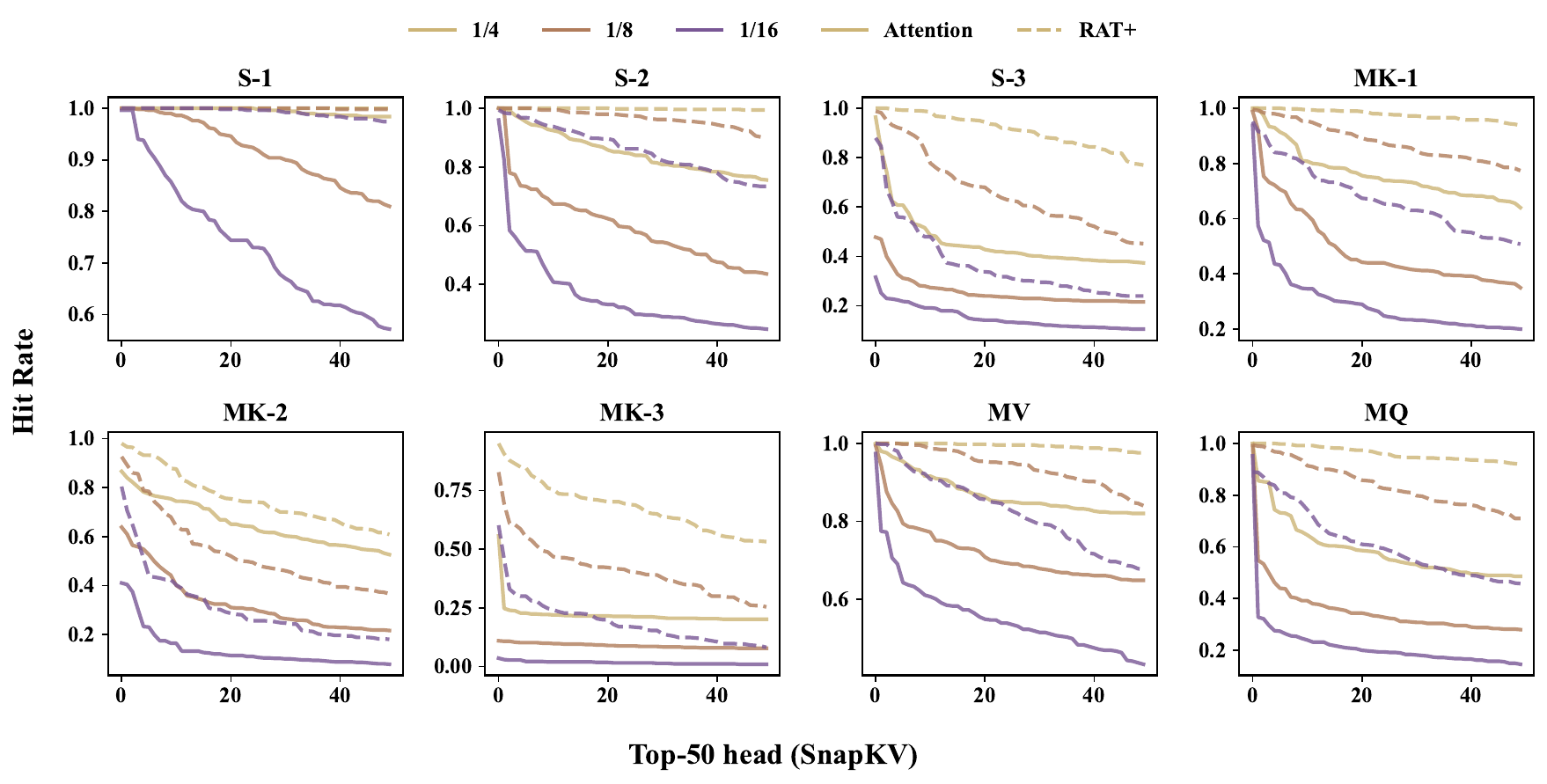}
    \caption{ Head-level hit-rate distributions for the top-50 heads ranked by hit rate for SnapKV.
    }
    \label{fig:hit_rate_snapkv}
\end{figure*}

Additional experimental results are provided in the appendix. We report more NIAH results in \autoref{tab:llama_7b} and \autoref{tab:olmo_7b}. We provide further analysis supporting H1 in \autoref{fig:hit_rate_snapkv} and \autoref{fig:hit_rate_quest}, and additional results supporting H2 in \autoref{tab:random_selector_full}.

Interestingly, for H2, we find that under random selection, SnapKV performs better than Quest, whereas it falls behind Quest under the original selectors. We conjecture that this is because SnapKV performs selection only once after prefilling, while Quest applies random selection at every decoding step, which introduces more dynamic noise throughout generation. 
\label{sec:supplementary_experiments}
\begin{table*}[htbp!]
\centering
\caption{Performance on eight NIAH tasks using the 7B checkpoints from \citet{ratplus}, where both the standard attention and RAT+ backbones are pretrained from scratch on 100B tokens. The budget denotes the fraction of KV entries accessed per token. Note that Quest and MoBA reduce KV access but still store the full KV cache. We also report results with a block size \(B\) of 128 for MoBA with better parallel efficiency, and a block size of 16 and number of block of 4 for Quest as an extreme setting. }
\label{tab:llama_7b}
\resizebox{\textwidth}{!}{
\begin{tabular}{lll ccc ccc cc}
\toprule
 \bf Backbone & \bf Top-\(K\) &  \bf Budget & \bf S-1 & \bf S-2 & \bf S-3 & \bf MK-1 & \bf MK-2 & \bf MK-3 & \bf MV & \bf MQ\\
\midrule
\bf Quest (\(B=64\))\\
\hspace{0.5em}Attention & \(K=16\) & \(1/4\) & 100	&100	&100	&100&	99.2	&45.2	&100	&100 \\
\hspace{0.5em}Attention & \(K=8\) &  \(1/8\) & 100 & 100 & 100 & 100 & 81.4 & 7.4 & 100 & 100 \\
\hspace{0.5em}Attention & \(K=4\) & \(1/16\) & 100 & 100 & 99.8 & 99.6 & 34.4 & 0.2 & 96.45 & 97.6 \\

\hspace{0.5em}RAT+ & \(K=16\)  & \(1/4\) & 100 & 100 & 100 & 99.8 & 100 & 94.2 & 100 & 100\\
\hspace{0.5em}RAT+ & \(K=8\)  &  \(1/8\) & 100 & 100 & 100 & 99.8 & 97.2 & 64.8 & 100 & 100\\
\hspace{0.5em}RAT+ & \(K=4\) & \(1/16\) & 100 & 100 & 99.6 & 99.6 & 70.4 & 18.6 & 99.45 & 99.5\\
\midrule
\bf Quest (\(B=16\)) \\
\hspace{0.5em}Attention & \(K=4\) & \(1/64\) & 100 & 100 & 90.6 & 99 & 49.6 & 0 & 94.65 & 94.5 \\
\hspace{0.5em}RAT+ & \(K=4\) & \(1/64\) & 100 & 100 & 95.4 & 99.4 & 92 & 26.2 & 98.6 & 98.2\\
\midrule
\bf MoBA (\(B=64\)) \\
\hspace{0.5em}Attention & \(K=16\) & \(1/4\) & 100 & 100 & 97.2 & 99.8 & 99 & 90.4 & 97.85 & 99.7 \\
\hspace{0.5em}Attention& \(K=8\) & \(1/8\) & 96 & 92.6 & 80 & 97 & 92.4 & 71.4 & 87.15 & 97.9 \\
\hspace{0.5em}Attention & \(K=4\) & \(1/16\) & 87 & 69.2 & 46 & 75.8 & 83.8 & 50 & 73.9 & 93.85 \\
\hspace{0.5em}RAT+ & \(K=16\)  & \(1/4\) &  100 & 100 & 100 & 99.8 & 100 & 99.4 & 100 & 100 \\
\hspace{0.5em}RAT+ & \(K=8\)  &  \(1/8\) & 100 & 100 & 100 & 99.8 & 100 & 99.2 & 99.8 & 99.95 \\
\hspace{0.5em}RAT+ & \(K=4\) & \(1/16\) & 100 & 100 & 100 & 99.8 & 100 & 96.2 & 98.35 & 99.85 \\
\midrule
\bf MoBA (\(B=128\)) \\
\hspace{0.5em}Attention & \(K=8\) & \(1/4\) & 97.8 & 98.4 & 83.6 & 98.2 & 97.2 & 77.2 & 93.8 & 98.9 \\
\hspace{0.5em}Attention & \(K=4\) & \(1/8\) & 82.8 & 71.4 & 46.4 & 80 & 86 & 57.4 & 77.3 & 94.5\\
\hspace{0.5em}RAT+ & \(K=8\) & \(1/4\) & 100 & 100 & 100 & 99.8 & 100 & 98 & 99.95 & 99.95  \\
\hspace{0.5em}RAT+ & \(K=4\) & \(1/8\) & 99.8 & 100 & 99.8 & 99.8 & 99.6 & 90.4 & 97.8 & 99.75\\
\midrule
\bf SnapKV \\
\hspace{0.5em}Attention & \(K=1024\) & \(1/4\) & 94.2 & 73.8 & 3 & 59.2 & 50.2 & 8.4 & 24.5 & 20.45 \\
\hspace{0.5em}Attention & \(K=512\) & \(1/8\) & 71.6 & 33.8 & 0.4 & 32.8 & 19.2 & 1.4 & 12.15 & 6.5 \\
\hspace{0.5em}Attention & \(K=256\) & \(1/16\)& 39.2 & 8.2 & 0.2 & 8.8 & 5.2 & 0 & 4.15 & 1.95 \\
\hspace{0.5em}RAT+ & \(K=1024\) & \(1/4\) & 97.2 & 97.8 & 39.6 & 94.4 & 88.2 & 18.8 & 82.8 & 87.85 \\
\hspace{0.5em}RAT+ & \(K=512\) & \(1/8\) & 93.8 & 91.6 & 24.2 & 85 & 69 & 4.6 & 60.75 & 69.15 \\
\hspace{0.5em}RAT+ & \(K=256\) & \(1/16\) & 84.2 & 73.6 & 15.8 & 70.4 & 41.2 & 0.4 & 42.9 & 50.35 \\
\bottomrule
\end{tabular}
}
\end{table*}

\begin{table*}[htbp!]
\centering

\caption{Performance on eight NIAH tasks using OLMo2-7B~\citep{olmo2}, which is trained on trillions of tokens. The RAT+ backbone is obtained through 10B-token continued pretraining with an additional exponentially decaying memory module. The budget denotes the fraction of KV entries accessed per token. Note that Quest and MoBA reduce KV access but still store the full KV cache. We also report results with a block size of 128 for MoBA for better parallel efficiency, and with a block size of 16 and four selected blocks for Quest as an extreme setting.}
\label{tab:olmo_7b}
\resizebox{\textwidth}{!}{
\begin{tabular}{lll ccc ccc cc}
\toprule
 \bf Backbone & \bf Top-\(K\) &  \bf Budget & \bf S-1 & \bf S-2 & \bf S-3 & \bf MK-1 & \bf MK-2 & \bf MK-3 & \bf MV & \bf MQ\\
\midrule
\bf Quest (\(B=64\))\\
\hspace{0.5em}Attention & \(K=16\) & \(1/4\) & 100 & 100 & 99.8 & 100 &     93.8 & 79.6 & 100 &     100 \\ 
\hspace{0.5em}Attention & \(K=8\) &  \(1/8\) & 100 & 100 & 99.8 & 99.2 & 87.2 & 28.2 & 98.55 & 99.55 \\
\hspace{0.5em}Attention & \(K=4\) & \(1/16\) & 100 & 99 & 97.4 & 76.4 & 68 & 1 & 63.65 & 82.8 \\ 
\hspace{0.5em}RAT+ & \(K=16\)  & \(1/4\) & 100 & 100 & 99.8 & 100 & 100 & 99.8 & 100 & 100 \\
\hspace{0.5em}RAT+ & \(K=8\)  &  \(1/8\) & 100 & 100 & 99.8 & 100 & 100 & 94.4 & 100 & 100 \\
\hspace{0.5em}RAT+ & \(K=4\) & \(1/16\) & 100 & 100 & 96.2 & 99.4 & 98.6 & 12.4 & 97.65 & 98.45\\
\midrule
\bf Quest (\(B=16\)) \\
\hspace{0.5em}Attention & \(K=4\) & \(1/64\) & 99.8 & 100 & 70.4 & 96.6 & 44.6 & 0 & 80.1 & 93.1\\
\hspace{0.5em}RAT+ & \(K=4\) & \(1/64\) & 100 & 100 & 79 & 100 & 99 & 3 & 99.9 & 99.3 \\

\midrule
\bf MoBA (\(B=64\)) \\
\hspace{0.5em}Attention & \(K=16\) & \(1/4\) &100 & 100 & 99.8 & 100 & 99.8 & 99.4 & 100 & 100 \\ 
\hspace{0.5em}Attention& \(K=8\) & \(1/8\) & 98.6 & 100 & 99.2 & 98.6 & 91.8 & 90.4 & 99.35 & 99.65 \\ 
\hspace{0.5em}Attention & \(K=4\) & \(1/16\) & 90.6 & 97.6 & 95.2 & 91.2 & 63.4 & 53.6 & 90.5 & 97.6 \\ 

\hspace{0.5em}RAT+ & \(K=16\)  & \(1/4\) & 100 & 100 & 99.8 & 100 & 99.8 & 99.8 & 100 & 100 \\
\hspace{0.5em}RAT+ & \(K=8\)  &  \(1/8\) & 100 & 100 & 99.6 & 100 & 98 & 99.6 & 100 & 100 \\
\hspace{0.5em}RAT+ & \(K=4\) & \(1/16\) & 99.8 & 100 & 99.6 & 99.4 & 85.6 & 94.8 & 98.4 & 99.85 \\

\midrule
\bf MoBA (\(B=128\)) \\
\hspace{0.5em}Attention & \(K=8\) & \(1/4\) & 99.8 & 100 & 99.8 & 99.2 & 95.6 & 94.6 & 99.6 & 99.6  \\
\hspace{0.5em}Attention & \(K=4\) & \(1/8\)         & 90 & 95.2 & 93.8 & 88.6 & 66 & 56.6 & 92.15 & 95.4 \\ 
\hspace{0.5em}RAT+ & \(K=8\) & \(1/4\) & 100 & 100 & 99.6 & 99.6 & 99 & 99.6 & 99.95 & 100\\
\hspace{0.5em}RAT+ & \(K=4\) & \(1/8\) & 99.6 & 100 & 99.2 & 98 & 82.2 & 92.4 & 98.05 & 99.45\\
\midrule
\bf SnapKV \\
\hspace{0.5em}Attention & \(K=1024\) & \(1/4\) & 95.4 & 100 & 28 & 100 & 100 & 0.8 & 82.3 & 100 \\
\hspace{0.5em}Attention & \(K=512\) & \(1/8\) & 84.2 & 99.8 & 4.2 & 99.6 & 100 & 0 & 61.35 & 98.55\\
\hspace{0.5em}Attention & \(K=256\) & \(1/16\)& 28.8 & 97 & 2.4 & 97.8 & 0.4 & 0 & 34.85 & 83.95 \\
\hspace{0.5em}RAT+ & \(K=1024\) & \(1/4\) & 100 & 100 & 77.6 & 100 & 100 & 98.4 & 99.9 & 100 \\
\hspace{0.5em}RAT+ & \(K=512\) & \(1/8\) & 100 & 100 & 35.8 & 100 & 100 & 84.6 & 96.05 & 100  \\
\hspace{0.5em}RAT+ & \(K=256\) & \(1/16\) & 100 & 100 & 3.2 & 100 & 100 & 32 & 65.25 & 100 \\
\bottomrule
\end{tabular}
}
\end{table*}

\begin{table*}[htbp!]
\centering
\caption{Random-selector performance under a \(1/4\) budget on OLMo checkpoints, reported as mean \(\pm\) standard deviation over five seeds. Quest and MoBA use a block size of 64 and select 16 blocks. SnapKV selects 1024 tokens. }
\label{tab:random_selector_full}
\resizebox{\linewidth}{!}{
\begin{tabular}{lll ccc ccc cc}
\toprule
 \bf Backbone & \bf S-1 & \bf S-2 & \bf S-3 & \bf MK-1 & \bf MK-2 & \bf MK-3 & \bf MV & \bf MQ\\
\midrule
\bf Quest \\
\hspace{0.5em}Attention &  48.36 ± 1.97 & 65.44 ± 1.25 & 0.88 ± 0.10 & 53.40 ± 0.67 & 63.28 ± 2.42 & 0.08 ± 0.10 & 39.57 ± 0.61 & 56.89 ± 1.20 \\ 
\hspace{0.5em}RAT+ & 83.80 ± 1.92 & 91.80 ± 1.60 & 30.00 ± 1.73 & 84.92 ± 0.73 & 91.52 ± 0.90 & 10.12 ± 1.16 & 74.40 ± 0.77 & 79.88 ± 0.31 \\ 
\midrule
\bf MoBA \\
\hspace{0.5em}Attention & 58.88 ± 1.84 & 91.12 ± 1.16 & 79.72 ± 1.63 & 73.64 ± 1.15 & 77.36 ± 2.97 & 65.08 ± 1.42 & 89.24 ± 0.65 & 94.66 ± 0.38 \\ 
\hspace{0.5em}RAT+ & 99.40 ± 0.13 & 100.00 ± 0.00 & 95.60 ± 0.63 & 94.84 ± 0.85 & 97.12 ± 0.81 & 85.72 ± 0.64 & 92.32 ± 0.32 & 98.85 ± 0.21 \\ 
\midrule
\bf SnapKV \\
\hspace{0.5em}Attention & 43.88 ± 1.54 & 63.24 ± 2.32 & 2.64 ± 0.08 & 63.16 ± 1.87 & 73.60 ± 1.96 & 0.56 ± 0.15 & 52.52 ± 0.98 & 60.63 ± 0.68 \\
\hspace{0.5em}RAT+ & 98.36 ± 0.66 & 99.56 ± 0.29 & 71.72 ± 1.37 & 99.16 ± 0.34 & 98.12 ± 0.70 & 35.56 ± 1.12 & 96.66 ± 0.35 & 98.93 ± 0.21 \\
\bottomrule
\end{tabular}
}
\end{table*}

\subsection{License information}

\begin{itemize}
    \item FineWeb-Edu (dataset): Open Data Commons License Attribution family. 

    Link: \url{https://huggingface.co/datasets/HuggingFaceFW/fineweb-edu}

    \item RULER benchmark (dataset): Apache 2.0 License. 
    
    Link: \url{https://github.com/NVIDIA/RULER}
    
    \item OLMo2-7B (model): Apache 2.0 License.
    
    Link: \url{https://huggingface.co/allenai/OLMo-2-1124-7B}
    
    \item RAT+ (model): Apache 2.0 License.
    
    Link: \url{https://huggingface.co/barpitf/ratplus}
    
\end{itemize}

\end{document}